\documentclass[runningheads]{llncs}

\usepackage{graphicx}
\usepackage[table]{xcolor}
\usepackage{booktabs}

\usepackage{amsthm}
\usepackage{xspace}
\usepackage{multirow}
\usepackage{amsmath}
\usepackage{float}
\usepackage{xcolor,colortbl}

\newcommand{\T}{\ensuremath{\mathcal{T}}}
\newcommand{\R}{\ensuremath{\mathcal{A}}}

\DeclareMathAlphabet{\pazocal}{OMS}{zplm}{m}{n}
\newcommand{\symL}{\pazocal{L}}
\newcommand{\symD}{\pazocal{D}}
\newcommand{\symS}{\pazocal{S}}

\begin{document}

\title{Aligning Biomedical Metadata with Ontologies Using Clustering and Embeddings}

\titlerunning{Aligning Biomedical Metadata with Ontologies}

\author{Rafael S.\ Gon\c{c}alves %\orcidID{0000-0003-1255-0125}
\and Maulik R.\ Kamdar %\orcidID{0000-0002-9898-0515}
\and Mark A.\ Musen %\orcidID{0000-0003-3325-793X}
}

\authorrunning{R. Gon\c{c}alves et al.}

\institute{Center for Biomedical Informatics Research\\Stanford University, CA, USA\\
\email{\{rafael.goncalves,maulikrk,musen\}@stanford.edu}}

\maketitle

%---------------- ABSTRACT ----------------%
\begin{abstract}
The metadata about scientific experiments published in online repositories have been shown to suffer from a high degree of representational heterogeneity---there are often many ways to represent the same type of information, such as a \emph{geographical location} via its latitude and longitude. To harness the potential that metadata have for discovering scientific data, it is crucial that they be represented in a uniform way that can be queried effectively. One step toward uniformly-represented metadata is to normalize the multiple, distinct field names used in metadata (e.g., \emph{lat lon}, \emph{lat and long}) to describe the same type of value. To that end, we present a new method based on clustering and embeddings (i.e., vector representations of words) to align metadata field names with ontology terms. We apply our method to biomedical metadata by generating embeddings for terms in biomedical ontologies from the BioPortal repository. We carried out a comparative study between our method and the NCBO Annotator, which revealed that our method yields more and substantially better alignments between metadata and ontology terms.

\end{abstract}

%---------------- INTRODUCTION ----------------%
\section{Introduction}
\label{sec:intro}

Metadata are crucial artifacts to facilitate the discovery and reuse of data. Publishers and funding agencies require scientists to share research data along with metadata. Sadly, the quality of metadata published in online repositories is typically poor. In previous work, we identified numerous anomalies in metadata hosted in the U.S. National Center for Biotechnology Information (NCBI) BioSample metadata repository \cite{Barrett2012BioProjectMetadata}, which describe biological samples used in biomedical experiments \cite{Goncalves2018TheExperiments}. For example, simple binary or numeric fields are often populated with inadequate values of different data types, and there are typically many ways to represent the same information in metadata fields (e.g., \emph{lat lon}, \emph{Lat-Long}, among dozens of other field names to represent a location via its latitude and longitude). This is not an idiosyncratic problem affecting just a single database---it is a pervasive issue that affects nearly all data repositories, and that is detrimental to the ability to effectively search and reuse data. 

%With the volume of discoveries that are powered by the high quantity of ever more discoverable linked datasets on the Web, science undoubtedly suffers as a consequence of the poor quality metadata that we observed and documented.

To make data Findable, Accessible, Interoperable, and Reusable (FAIR) \cite{Wilkinson2016}, it is necessary that the metadata that describe the data be of high quality. One step toward generating better quality metadata out of legacy metadata is to normalize the multiple field names that are used interchangeably in metadata to describe the same thing. By doing so, we can reduce the representational heterogeneity that currently hinders metadata repositories and thus improve the searchability of the associated data. In our study of the quality of BioSample metadata, we identified that metadata authors tend to use off-the-cuff field names to annotate their datasets even when there are standardized terms (specified by the repositories) to use for a particular type of value. Metadata field names are rarely linked to terms from ontologies or metadata vocabularies. While the use of ontology terms is encouraged in some fields, there is no mechanism to enforce or verify this suggestion. Ideally, the field names used in a metadata repository should be drawn from ontologies, and there should be one single field name, rather than many, to describe each distinct type of metadata value.

In this paper, we present a new method to semi-automatically align arbitrary strings with ontology terms using \emph{embeddings}. Embeddings are representations of words or phrases (or other types of entities) as low-dimensional numeric vectors. Word embeddings are capable of capturing contextual information and relations between different words in a text corpus \cite{goldberg2014word2vec,pennington2014glove}. Our method works by clustering input strings according to a string distance metric, such as the Levenshtein edit distance, and then comparing embeddings for the input strings with embeddings computed for the human-readable labels of ontology terms. The term alignments are ranked by taking into account the cosine distance between embeddings, and the edit distance between input strings and ontology term labels. To show the efficacy of our method, we align a corpus of metadata field names taken from the BioSample metadata repository with terms from ontologies in BioPortal \cite{Noy2009BioPortal:Mouse}. We compare three clustering methods over six different distance metrics to determine which combination works best for the BioSample corpus. Then, we carry out a comparative study of the alignments found by our method with those found by the NCBO Annotator \cite{Jonquet2009-ds}. Finally, we conduct semi-structured interviews with subject-matter experts to verify: (a) the quality of our alignments compared to those that the NCBO Annotator finds, and (b) the appropriateness of our alignments when the NCBO Annotator finds none.

%---------------- RELATED WORK ----------------%
\section{Related Work}
\label{sec:related}

The NCBO Annotator is a reference service for annotating biomedical data and text with terms from biomedical ontologies. The service works directly with ontologies hosted in the BioPortal ontology repository. The NCBO Annotator relies on the Mgrep concept recognizer, developed at the University of Michigan, to match arbitrary text against a set of dictionary terms provided by BioPortal ontologies \cite{shah2009comparison}. The NCBO Annotator additionally exploits is-a relations between terms to expand the annotations found with Mgrep and to rank the alignments. 

%Associated with the NCBO Annotator is the NCBO Recommender \cite{martinez2017ncbo}, a service that takes text as input and finds a set of ontologies that are appropriate to annotate the text, according to various criteria such as how many words in the text can be aligned with ontology terms. The NCBO Recommender uses the NCBO Annotator to find alignments between individual words or terms in the input text and ontology terms. 

%The work by McCrae targets alignment between structured entities, while in our work we align unstructured text with structured entities.  For our task, the most comparable works are the NCBO-NCBO Annotator and a similar tool called Zooma, hosted at EBI-OLS.  MetaMap is also related, although only applicable to UMLS concepts.  The NCBO-NCBO Annotator has been the reference tool for annotating unstructured biomedical text with ontology terms, so an in-depth comparison with this method seems most fitting.

There are several recent methods to generate embeddings for entities in an RDF graph or OWL ontology. These embeddings, often called knowledge graph embeddings, require triples of the form $\langle$\textit{head entity, relation, tail entity}$\rangle$. Translation-based methods represent each $head$ or $tail$ entity as a point in a vector space, and the $relation$ represents a vector translation function in a hyperplane (i.e., the $head$ entity vector can be translated to the $tail$ entity vector, using a geometrical function over the $relation$ vector) \cite{wang2014knowledge,lin2015learning,socher2013reasoning}. Other methods to generate these knowledge graph embeddings are inspired by language modeling approaches, which rely on sequences of words in a text corpus---that is, random walks are carried out on an RDF graph to generate sequences of entities in nearby proximity, and these sequences can be used to generate latent numerical representations of entities \cite{ristoski2016rdf2vec}. While these methods have been effective for link prediction in knowledge graphs, triple classification, and fact extraction, they cannot be used to assess semantic similarity of different terms by examining how they are used in the context of literature. 

Word embeddings can be generated from a corpus of textual documents through popular approaches such as skip-gram and continuous bag of words models \cite{pennington2014glove,goldberg2014word2vec}. Word embeddings generated from a corpus of biomedical abstracts have been shown to aid in the discovery of novel drug--reaction relations \cite{percha2018global}. Embeddings can be computed for ontology term labels by using a weighted mean of word embeddings contained within the label. Such term embeddings have been used to align elements in the vocabularies of biomedical RDF graphs \cite{kamdaremperical}. Such an approach does not require re-training the term embeddings (a shortcoming of phrase embedding approaches \cite{passos2014lexicon}), it can use domain-specific context (e.g., the biomedical significance behind the term \textit{tumor region}, and similarity to other terms such as \textit{cell region} or \textit{site of cancer}), and it can enable similarity between terms with arbitrary word lengths.

The existing works to align biomedical text with ontology terms have the following shortcomings: they do not use vector space embeddings of terms \cite{jimenez2011logmap}; they generate ontology term embeddings using the structure of the ontology rather than background knowledge from textual data \cite{smaili2018opa2vec}; and they perform alignments against only a few ontologies containing a small number of terms. Conventional syntactic similarity metrics (e.g., Levenshtein distance) do not capture the context where terms are used. Some techniques use embeddings computed with background knowledge from Wikipedia or other generic corpora. Such embeddings are not as appropriate for our case, because biomedical metadata field names consist of multi-word, domain-specific, and complex terms, which are not typically defined or mentioned together in generic corpora \cite{wang2018comparison}. Using biomedical background knowledge is essential to improve the efficacy of alignments, since most terms are biomedicine-specific; and to provide textual contexts where the aligned terms are used, which is essential to help experts verify alignments.

%---------------- METHODS ----------------%
\section{Methods}
\label{sec:methods}

\begin{figure}[ht!]
\centering
\includegraphics[scale=0.37]{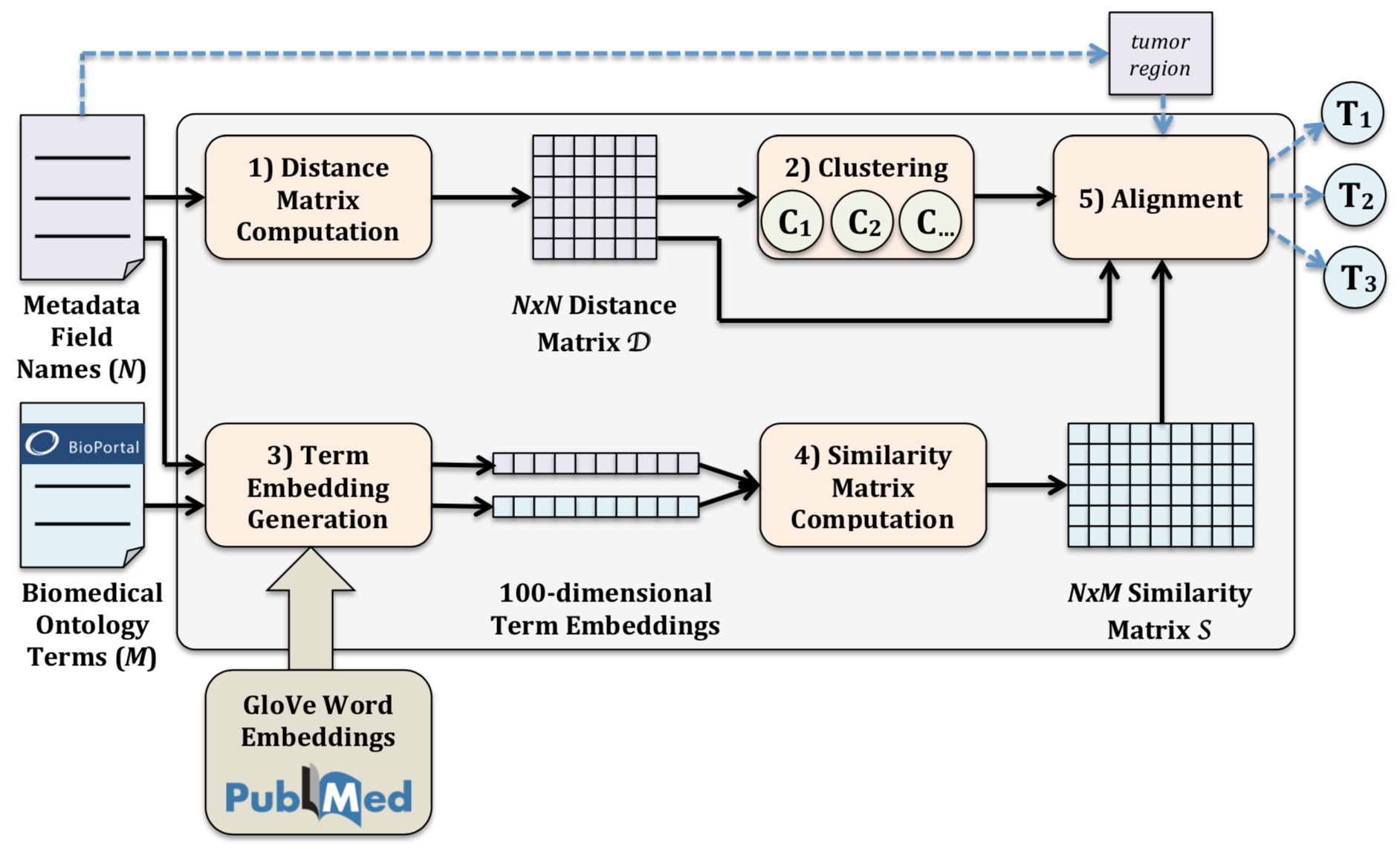}
\vspace{-5.5mm}\caption[Components of our metadata alignment method]{\textbf{Components of our metadata alignment method.} In this figure, methods are colored in orange, metadata-related variables are colored in purple, and ontology-related variables are colored in blue. \textbf{(1) Distance Matrix Computation:} Our method uses a string distance metric to compare $N$ metadata field names with each other, generating a distance matrix $\symD$ of shape $N\times N$. \textbf{(2) Clustering:} A clustering function takes as input the distance matrix $\symD$ and generates clusters of metadata field names. \textbf{(3) Term Embedding Generation:} Metadata field names and terms from biomedical ontologies are represented in a 100-dimensional vector space, using word embeddings generated from PubMed (\url{https://www.ncbi.nlm.nih.gov/pubmed}) abstracts. \textbf{(4) Similarity Matrix Computation:} Our method computes cosine similarity scores between term embeddings of $N$ metadata field names and $M$ ontology terms, generating an $N\times M$ similarity matrix $\symS$. \textbf{(5) Alignment:} Given a metadata field name (e.g., \textit{tumor region}), our method selects the cluster containing the field name, and then chooses the top metadata field names that have a minimal distance (using $\symD$) from the given metadata field name. Using these highly similar field names, the distance matrix $\symD$, and the similarity matrix $\symS$, our method generates and ranks ontology alignments $\{T_1, T_2, T_3, \ldots\}$, shown on the right side of the figure as output.}
\label{fig:approach}
\end{figure}

The approach we designed to align a collection of arbitrary strings with ontology terms, shown in Figure \ref{fig:approach}, consists of the following steps. (1) We cluster input strings according to a distance metric (e.g., Levenshtein distance), which provides a grouping of syntactically similar strings. (2) We compare embeddings of input strings with ontology term embeddings trained on a corpus of biomedical ontologies. (3) We generate alignments between input strings and ontology terms by attending to the distance between the strings and the ontology terms.

\vspace{-4.5mm}
\vspace{-4mm}

For the purpose of human analysis, an ideal combination of a clustering function and a distance metric would yield relatively small clusters, for example, no bigger than 100 elements. We can test this automatically, by attending to observations such as the average size of the clusters, the size of the biggest cluster, the cluster size variance, and so on. For the purpose of providing a reliable basis upon which to compute alignments, an ideal combination would generate clusters whose elements are so syntactically or semantically close to one another that they are likely to represent related aspects of the data. This is non-trivial to verify in any automated setting. We propose a proxy for measuring how good a set of clusters is for an alignment task such as ours, which can be automatically tested. Given an alignment function, we determine what is the maximum number of terms that can be aligned with a single ontology per cluster. Having at least one alignment for each cluster gives us a candidate alignment that we can use to inform alignments for other elements in the same cluster.

\vspace{-3mm}
%---------------- DISTANCE METRICS ----------------%
\subsubsection{1) Distance Matrix Computation}
\label{sec:clustering}

There are various string distance metrics shown to be useful in practice. For example, Levenshtein edit distance is widely deployed in word processors to detect typos. In our selection of distance metrics, we included the Levenshtein distance metric along with its variant, Damerau-Levenshtein.  We tested cosine distance as it is the basis for our semantic alignment. We also tested Jaro distance, its variant Jaro-Winkler, and Jaccard distance (set-based, with exact match), because they are well-known and widely used metrics. We use a distance metric to compare $N$ metadata field names with each other, thus generating a distance matrix $\symD$ of shape $N\times N$.

% We used the \emph{jellyfish} (\url{https://github.com/jamesturk/jellyfish}) and the \emph{nltk} (\url{https://www.nltk.org/}) Python packages for computing the pairwise distances between strings. 

\vspace{-3mm}
%---------------- CLUSTERING ----------------%
\subsubsection{2) Clustering}
\label{sec:clustering}

Clustering functions applicable to our method must not require an upfront specification of the number of clusters. This is intuitively necessary since we have no a priori knowledge of what the data are. When selecting clustering functions, we had prior knowledge that affinity propagation \cite{Frey2007ClusteringPoints} worked well to cluster biomedical terms according to Levenshtein distance (see \cite{Goncalves2018TheExperiments}). We compared affinity propagation with a highly common density-based algorithm--Density-Based Spatial Clustering of Applications with Noise (DBSCAN) \cite{ester1996density}, and with Hierarchical DBSCAN (HDBSCAN) \cite{mcinnes2017hdbscan}, which is a hierarchical type of clustering that builds on DBSCAN. Each clustering function takes as input a distance matrix and assigns metadata field names to disjoint clusters. 

%We used the implementation of the DBSCAN and affinity propagation algorithms in the \emph{scikit-learn} Python package (\url{https://scikit-learn.org}), and the implementation of the HDBSCAN in a \emph{skikit-learn} compatible package (\url{https://github.com/scikit-learn-contrib/hdbscan}).

\vspace{-3mm}
%---------------- EMBEDDING GENERATION ----------------%
\subsubsection{3) Term Embedding Generation}
\label{sec:training}

We describe the method we designed to represent metadata field names and ontology terms as high-dimensional numerical vectors, henceforth called \emph{term embeddings}, based on the GloVe (Global Vectors for Word Representation) algorithm \cite{pennington2014glove}. We generated word embeddings using a corpus of approximately 30 million PubMed abstracts from the MEDLINE database--a database of journal citations and abstracts for biomedical literature \cite{koster2007parsing}. For this work, we used only the title and the abstract of each citation. We performed the same preprocessing steps to process biomedical publication abstracts (i.e., tokenization, medical entity normalization) from the PubMed repository as Percha, et al. \cite{percha2018global}.

Word embeddings represent each word as a high-dimensional numerical vector based on co-occurrence counts of those words as observed in publications. We used the GloVe algorithm to generate 100-dimensional word embeddings from the tokenized biomedical publications. The vectors are generated after a training phase of 20 iterations with an $\alpha$ learning rate of 0.75. These parameters (e.g., vector dimensions) are inspired by Percha et al. \cite{percha2018global}, who successfully used and evaluated biomedical word embeddings to discover novel drug--drug interactions.

%100-dimensional word embedding vectors and inverse document frequency (IDF) statistics are generated for a vocabulary of 2,531,989 words from approximately 30 million PubMed biomedical publication abstracts. Using these word embedding vectors and IDF statistics, 100-dimensional URI embedding vectors are generated for different schema elements URIs in the

We represent each metadata field name or ontology term in a high dimensional space using word embedding vectors. We generate a vocabulary of 2,531,989 words from 30 million PubMed biomedical publication abstracts. The words in our vocabulary appear in at least 5 distinct abstracts. We determined this threshold empirically, by identifying words that might be important for metadata field names (e.g., an unusual term such as \texttt{zymosterone}, which is a compound involved in the synthesis of cholesterol) and that are mentioned in only 5 abstracts. Subsequently, we represent each word in a 100-dimensional numerical space called the word embedding vector. We generate a term embedding vector by computing a weighted average of the words in the term label, with the weights being the Inverse-Document-Frequency (IDF) statistic for each word. We created a default embedding vector and IDF statistic (0.01) for any word not in the vocabulary. We show the equation to generate a term embedding vector below. 
        \begin{align}
            \mathbf{x}_{(term)} = \frac{\sum_{w_i \in \symL(term)}idf_{(w_i)}*\mathbf{x}_{(w_i)}}{\sum_{w_i \in \symL(term)}idf_{(w_i)}} 
        \end{align}
Here, $\mathbf{x}_{(w_i)}$ represents the 100-dimensional word embedding vector, and $\symL(term)$ is the term label. For ontology terms, the term label $\symL(term)$ is extracted from the values of annotations properties commonly used in biomedical ontologies to encode human-readable labels (i.e. \emph{rdfs:label} and \emph{skos:prefLabel}), whereas for metadata field names, the term label is the field name.

\vspace{-3mm}
%---------------- SIMILARITY MATRIX COMPUTATION ----------------%
\subsubsection{4) Similarity Matrix Computation}

We compute a cosine similarity matrix $\symS$ for the $N$ metadata field names and $M$ ontology terms. Hence, the shape of matrix $\symS$ is $N\times M$, and an individual cell in this matrix $\symS_{ij}$ is the cosine similarity score $co\text{-}sim(N_i, M_j)$ between the metadata field name $N_i$ and the ontology term $M_j$. We present the equation to compute $co\text{-}sim (N_i, M_j)$ below.
\begin{align}
    co\text{-}sim (N_i, M_j) = \frac{\mathbf{x}_{(N_i)}\cdot\mathbf{x}_{(M_j)}}{||\mathbf{x}_{(N_i)}||||\mathbf{x}_{(M_j)}||} = \frac{\sum_{k=1}^{n=100}x_{(N_i)k}*x_{(M_j)k}}{\sqrt{\sum_{k=1}^{n=100}x_{(N_i)k}^2}\sqrt{\sum_{k=1}^{n=100}x_{(M_j)k}^2}}
\end{align}
In the above equation, $\mathbf{x}_{(N_i)}$ and $\mathbf{x}_{(M_j)}$ are the 100-dimensional term embedding vectors for the metadata field name $N_i$ and the ontology term $M_j$ respectively.

\vspace{-3mm}
%---------------- ALIGNMENT ----------------%
\subsubsection{5) Alignment}

We specify our approach to find alignments between strings and ontology terms in Definition \ref{def:alignfunct}.

\begin{definition} Let \emph{align$_r$(S, C$_\Psi$, D$_m$)} be a function that takes a set of strings \emph{S}, a collection of clusters \emph{C} computed according to a clustering function~$\Psi$, and a matrix \emph{D} of the pairwise distances between all strings in \emph{S} computed according to a distance metric \emph{m}, and returns an alignment map \emph{\R~:=~S~$\rightarrow$~\T} that maps each input string in \emph{S} to a list of recommended ontology terms \T. We say a string $s \in S$ is aligned with an ontology term $t \in \T$ if the average of the cosine similarity between the embeddings of \emph{s} and \emph{t}, denoted $w(s)$ and $w(t)$ respectively, and the edit similarity between \emph{s} and \emph{t} is above a threshold \emph{r}, as follows.\\

$\R~:=~\{s \rightarrow t \mid s \in S \wedge \frac{\emph{co-sim}(w(s), w(t)) + \emph{edit-sim}(s,t)}{2} > r \}$
\label{def:alignfunct}
\end{definition}
\noindent 
We use both the cosine and edit similarities between strings to prevent false positive alignments between completely unrelated strings, which could result from aligning using cosine similarity alone. We have anecdotal evidence that the IDF ranking sometimes unduly attributes too much weight to certain words in metadata field names and throws off alignments. On the other hand, by taking both measures into account we run the danger of missing alignments between strings that are syntactically very different but semantically equivalent. We experimented with different weighting schemes between cosine and edit distance, and found the best compromise to be the average between \emph{co-sim} and \emph{edit-sim}. 

\vspace{-3mm}
%---------------- MATERIALS ----------------%
\subsubsection{Materials}
\label{sec:materials}

The metadata in the NCBI BioSample are specified through fields that take the form of \emph{name--value pairs}, for example, \textbf{geo location:} \emph{Alaska}. Users of BioSample can use standard \emph{field names} specified by BioSample, or they can (and often do) coin new names for their fields. We gathered all the fields names used across all metadata in BioSample. There are a total of 18,198 syntactically distinct field names in the BioSample metadata. We then normalized the metadata field names by replacing all non-alphabetic symbols with spaces, splitting words concatenated in CamelCase notation, converting all strings to lower case, and trimming all but one space between words. After removing duplicates, and strings which had fewer than 3 characters that we considered to be inappropriately short as metadata field names, we ended up with 15,553 unique field names. To generate embeddings, we used a corpus of ontologies from BioPortal composed of 675 ontologies, which was extracted on May 25, 2018.

%---------------- RESULTS ----------------%
\section{Results}
\label{sec:results}

In this section we present the results of clustering and aligning the metadata field names from the BioSample repository with terms from ontologies in the BioPortal repository. We first carried out an exploratory study of clustering functions and distance metrics in order to identify a suitable combination for the BioSample corpus. Based on this experiment, described in Section \ref{sec:clustresults}, we hoped to draw a theory about selecting an ideal combination of a clustering function and distance metric that can be automated in the future. Subsequently, we performed a comparative study of the alignments found by our method and those found by the NCBO Annotator. We compare the two methods according to whether they found at least one alignment per cluster, and according to how many field names they were able to align. We describe this study in Section \ref{sec:alignresults}.

%---------------- CLUSTERING RESULTS ----------------%
\subsection{Clustering Metadata Field Names}
\label{sec:clustresults}

In this experiment, we normalized the BioSample field names as described in Section \ref{sec:materials}, and then we evaluated combinations of clustering methods and distance metrics according to the following measures: number of clusters generated; average, median, and standard deviation of the number of field names in each cluster; number of field names in the biggest and smallest cluster; and number of clusters of the smallest size. We show the results in Table \ref{tab:clustresultstbl}. 

\begin{table}[t]
\centering
\caption{\textbf{Results of clustering metadata field names}. The table shows, from left to right, for each combination of clustering function and distance metric: the number of clusters; the average cluster size; the median cluster size; the standard deviation of the cluster size; the size of the biggest cluster; the size of the smallest cluster; and the number of clusters of the smallest size. The best results are shown in bold.}
\label{tab:clustresultstbl}
\begin{tabular}{l l l l l l l l l}
\toprule
Clustering & Distance & Nr.\ of & Avg & Median & Std & Biggest & Smallest & Nr.\ clusters of\\
method & metric & clusters & size & size & dev.\ & cluster & cluster & smallest size\\\midrule
\multirow{6}{*}{HDBSCAN} & Levenshtein & 641 & 27 & 10 & 94 & 1671 & 2 & 140\\
 & Damerau & 712 & 24 & 8 & 85 & 1671 & 2 & 174\\
 & Jaro & 871 & 20 & 7 & 91 & 2232 & 2 & 230\\
 & Jaro-Winkler & 1053 & 16 & 6 & 80 & 2385 & 2 & 255\\
 & Jaccard & 721 & 24 & 9 & 81 & 1762 & 2 & 178\\
 & Cosine & 1381 & 12 & 6 & 89 & 3091 & 2 & 314\\\midrule
\multirow{6}{*}{DBSCAN} & Levenshtein & 1191 & 14 & 2 & 279 & 8099 & 2 & 805\\
 & Damerau & 1178 & 14 & 2 & 280 & 8037 & 2 & 795\\
 & Jaro & 1623 & 10 & 2 & 192 & 7006 & 2 & 924\\
 & Jaro-Winkler & 941 & 18 & 2 & 345 & 10151 & 2 & 514\\
 & Jaccard & 1045 & 16 & 2 & 306 & 7660 & 2 & 755\\
 & Cosine & 1150 & 15 & 2 & 258 & 6844 & 2 & 702\\\midrule
\multirow{6}{*}{\textbf{AP}} & Levenshtein & 2145 & 8 & 5 & 8 & 60 & 1 & 491\\
 & Damerau & 2146 & 8 & 5 & 8 & 58 & 1 & 495\\
 & Jaro & 1495 & 11 & 10 & 7 & 50 & 1 & 1\\
 & \textbf{Jaro-Winkler} & \textbf{1387} & \textbf{12} & \textbf{10} & \textbf{9} & \textbf{61} & \textbf{2} & \textbf{13}\\
 & Jaccard & 1743 & 10 & 8 & 7 & 43 & 1 & 264\\
 & Cosine & 1415 & 12 & 7 & 32 & 1126 & 1 & 159\\
\bottomrule
\end{tabular}
\vspace{-4mm}
\end{table}

The HDBSCAN algorithm generated clusters that were of reasonable size for human analysis. The average cluster size across the board was around 20 field names, and the median cluster size was 10 or fewer. However, every distance metric we used resulted in at least one huge cluster with over 1,500 field names, and hundreds of tiny clusters with only 2 names each. Overall the clusters generated using HDBSCAN were not particularly suitable for human analysis. 

The results of clustering metadata field names using DBSCAN were very poor with all distance metrics. This method yielded extremely large clusters containing around 8,000 field names and over 500 clusters with only 2 field names. There is significant variability in the size of the clusters too, as illustrated by the high standard deviation of the size of clusters. Generally, the clusters generated using DBSCAN are even less suitable for human analysis than HDBSCAN. 

Clustering metadata field names using affinity propagation resulted in seemingly high quality clusters in terms of size. There were on average 1,783 clusters across all distance metrics, which is higher than the other clustering methods. However, this means that the clusters are generally much smaller. On average there were 10 field names per cluster, and the median size was 8 field names. The sizes of the large clusters found by affinity propagation were especially encouraging; the largest cluster was found using the Jaro-Winkler distance metric and it has 61 field names. In comparison with the other methods, this is highly encouraging. The number of clusters of the smallest size varies between distance metrics. Clustering based on the Levenshtein, Damerau, or Jaccard distances results in many clusters with only one field name. Using the Jaro-Winkler distance as a basis for clustering seems to yield the best overall results---it results in a desirable number of metadata field names per cluster, it has low variability of cluster size, and it results in neither too big clusters nor too small clusters.

%---------------- ALIGNMENT RESULTS  ----------------%
\subsection{Alignments Between Field Names and Ontology Terms}
\label{sec:alignresults}

In this experiment, we aimed to verify how well our approach can align clusters and their elements with ontologies and ontology terms, using different combinations of clustering methods and distance metrics. We compare the alignments obtained using our approach with those obtained using the NCBO Annotator. For each cluster in a set of clusters, we set out to find one ontology that provides the most alignments for the elements in that cluster---that is, an \emph{ontology recommendation} for a cluster---and we record the number of field names aligned with that ontology. The alignments of our method are based on a minimum cosine similarity (threshold $r$ in Definition \ref{def:alignfunct}) of 0.85 between the embeddings of field names and ontology terms. The results of this experiment are in Table \ref{tab:alignresultstbl}.

\definecolor{Gray}{gray}{0.93}
\newcolumntype{a}{>{\columncolor{Gray}}l}

\begin{table}[t]
\centering
\setlength{\tabcolsep}{3pt}
\caption{\textbf{Comparison of alignments found by our approach and by the NCBO Annotator}. The table shows for each combination of clustering method and distance metric: the number of ontology recommendations obtained for all clusters (``Nr.\ recs.''); the percentage of clusters that have one ontology recommendation, i.e., coverage (``Cov.''); the average/median number of field names in each cluster that were aligned with ontology terms. The best results are shown in bold.}
\label{tab:alignresultstbl}
\begin{tabular}{l l | l l l | l l l}
%\cmidrule{3-8}
& & \multicolumn{3}{l}{\textbf{Our Approach}} & \multicolumn{3}{|l}{\textbf{NCBO Annotator}} \\
\toprule
Clustering & Distance & Nr.\ & Cov.\ & Avg./median & Nr.\ & Cov.\ & Avg./median \\
method & metric & recs.\ & & fields covered & recs.\ & & fields covered\\
\toprule
\multirow{6}{*}{HDBSCAN} & Levenshtein & 625 & 98\% & 11 / 5 & 390 & 61\% & 3 / 1\\
 & Damerau & 695 & 98\% & 10 / 4 & 428 & 60\% & 3 / 2\\
 & Jaro & 853 & 98\% & 9 / 4 & 403 & 46\% & 3 / 2\\
 & Jaro-Winkler & 1027 & 98\% & 8 / 4 & 678 & 64\% & 2 / 1\\
 & Jaccard & 711 & 99\% & 10 / 4 & 441 & 61\% & 3 / 1\\
 & Cosine & 1355 & 98\% & 6 / 4 & 859 & 62\% & 2 / 1\\\midrule
\multirow{6}{*}{DBSCAN} & Levenshtein & 1042 & 87\% & 7 / 2 & 309 & 26\% & 1 / 1\\
 & Damerau & 1031 & 88\% & 7 / 2 & 306 & 26\% & 2 / 1\\
 & Jaro & 1448 & 89\% & 6 / 2 & 642 & 40\% & 2 / 1\\
 & Jaro-Winkler & 821 & 87\% & 9 / 2 & 358 & 38\% & 2 / 1\\
 & Jaccard & 937 & 90\% & 7 / 2 & 375 & 36\% & 1 / 1\\
 & Cosine & 1037 & 90\% & 7 / 2 & 440 & 38\% & 1 / 1\\\midrule
\multirow{6}{*}{\textbf{AP}} & Levenshtein & 1962 & 91\% & 5 / 4 & 773 & 36\% & 3 / 1\\
 & Damerau & 1961 & 91\% & 5 / 4 & 774 & 36\% & 3 / 1\\
 & Jaro & 1432 & 96\% & 7 / 6 & 867 & 58\% & 2 / 2\\
 & \textbf{Jaro-Winkler} & \textbf{1295} & \textbf{93\%} & \textbf{8 / 7} & \textbf{952} & \textbf{69\%} & \textbf{2 / 2}\\
 & Jaccard & 1609 & 92\% & 6 / 5 & 1066 & 61\% & 2 / 2\\
 & Cosine & 1356 & 96\% & 7 / 5 & 913 & 65\% & 2 / 2\\
\bottomrule
\end{tabular}
\end{table}

Our method found an ontology with which to align cluster elements for over 90\% of clusters, on average, across all combinations of clustering methods and distance metrics. On the other hand, the NCBO Annotator only found ontologies for less than 50\% of clusters on average. The best coverage obtained using the NCBO Annotator is using affinity propagation and Jaro-Winkler distance, for which 69\% of the clusters were aligned with ontologies. While this combination is not the best that our method found, if we take into account the average coverage of both our method and the NCBO Annotator, the combination just mentioned is still the best one, yielding 81\% average coverage for both methods.

Using our method, we aligned 7 metadata field names per cluster with ontology terms across all combinations in Table \ref{tab:alignresultstbl}, while the NCBO Annotator only aligned 2 field names, on average. Taking into account that the mean cluster size is 12 elements using affinity propagation and Jaro-Winkler, aligning 8 of those elements on average is a success. The most metadata field names aligned per cluster was achieved using the HDBSCAN algorithm. This is unsurprising, since HDBSCAN produces much larger clusters, and so it is more likely that it can find many more alignments per cluster and thus have better overall coverage.

Overall, using the NCBO Annotator we found alignments for 12,454 metadata field names, while using our method we found alignments for all terms. The average similarity score of the topmost alignments for all field names is 0.94 (where 1 means an exact match), which looks highly promising.

%---------------- EVALUATION BY EXPERT PANEL ----------------%
\section{Evaluation by Expert Panel}
\label{sec:panelevaluation}

To investigate the quality of the alignments found by our method, we designed an evaluation with a panel of subject-matter experts to verify the following two hypotheses. \emph{(H1)} The ontology term alignments provided by our method are preferable to alignments provided by the NCBO Annotator for the same metadata field name. \emph{(H2)} The ontology term alignments provided by our method are adequate to describe metadata field names even when the NCBO Annotator does not provide alignments for those field names. We continued to use a minimum similarity score (threshold $r$ in Definition \ref{def:alignfunct}) of 0.85 between BioSample metadata field name and ontology terms, to ensure that our method provided good quality alignments.

%---------------- INTERVIEW DESIGN ----------------%
\subsection{Semi-Structured Interview Design}
\label{sec:surveydesign}

We designed a semi-structured interview format to test our hypotheses, and we conducted interviews with four experts who are biomedical metadata experts. Our panel experts were specifically selected as they all have extensive experience in: engineering and working with metadata authoring and publishing software; constructing metadata forms for specific datasets and community standards; using ontologies and metadata vocabularies to annotate data; and working with users and developers to build infrastructure for describing scientific data. All experts have backgrounds and degrees in computer science.

We compiled one list of metadata field names to test each hypothesis. \emph{List 1} to test \emph{H1} was composed of 6 metadata field names randomly drawn from the set of metadata field names for which both our method and the NCBO Annotator found alignments. \emph{List 2} to test \emph{H2} was composed of 6 metadata field names randomly drawn from the set of metadata field names for which the NCBO Annotator did not find alignments. The selected metadata field names for the expert evaluation are listed in Table \ref{tab:metadatafieldnames}. The questions for each expert were randomly selected from List 1 and 2 in such a way that, for every metadata field name, we would have 3 responses from the experts.

\begin{table}[t]
\centering
\setlength{\tabcolsep}{8pt}
\caption{\textbf{Lists of metadata field names used in the interviews.} The table shows the metadata field names selected for \emph{List 1} to test \emph{H1}, and for \emph{List 2} to test \emph{H2}.}
\label{tab:metadatafieldnames}
\begin{tabular}{r l r l}
\toprule
\multicolumn{2}{l}{\textbf{List 1}} & \multicolumn{2}{l}{\textbf{List 2}} \\
\midrule
Q1 & sample depth m & Q7 & scientif name\\
Q2 & mapped reference genome & Q8 & patientnumber\\
Q3 & tissue source & Q9 & partecipants\\
Q4 & isolate id & Q10 & stimuli\\
Q5 & tumor region & Q11 & cryptophytes\\
Q6 & cardiovascular failure & Q12 & relhumidityavg\\
\bottomrule
\end{tabular}
\end{table}

In our study we asked the experts to discuss if and how each metadata field name is related to the ontology terms with which it is aligned. We use qualitative categories, which we describe in Table \ref{tab:responsecategories}, to gather feedback from experts. We chose this style of scale since a typical Likert-style scale is unlikely to be informative as to precisely how a given term relates to a field name (according to the experts' understanding and interpretation of the text). For example, it is preferable to know, for a given metadata field name \emph{bird species}, that the ontology term \emph{species} is \textbf{more general} than \emph{bird species}, or that \emph{owl} is \textbf{more specific}, than it is to know that an ontology term is more or less appropriate to describe, or better or worse aligned with some field name.

\begin{table}[t]
\centering
\setlength{\tabcolsep}{8pt}
\caption{\textbf{Expert response categories used in the interviews.} The table shows the categories that we used to gather feedback about how metadata field names related to the recommended ontology terms. In the second column we show example recommended ontology terms for the input field name \emph{``bird species''}.}
\label{tab:responsecategories}
\begin{tabular}{l l}
\toprule
\textbf{Category} & \textbf{Examples for \emph{``bird species''}} \\
\midrule
Unrelated & mouse\\
Looks unrelated & bird by species\\
May be related & bird species identifier\\
More general & species\\
More specific & owl\\
Looks similar & bird species name\\
Identical & species of bird\\
\bottomrule
\end{tabular}
\vspace{-4mm}
\end{table}

The format of the interview was as follows:

\begin{enumerate}
    \item We described the purpose of the study---to evaluate different algorithms for aligning metadata field names with ontology terms. 
    \item We described the challenge that our method meets---to normalize the many different field names used in metadata records to describe the same data.
    \item We listed typical questions that experts could ask of the interviewer---to give example values for the metadata fields; to clarify the task or the questions; to clarify or give examples of the categories in Table \ref{tab:responsecategories}. 
    \item We described the task that the experts were about to carry out---for each question out of 9 questions:
    \begin{enumerate}
        \item We showed the metadata field name, and asked the expert to describe her/his understanding of the field name. We ranked their understanding according to whether the meaning of the field name was \emph{clear}, \emph{roughly clear}, or \emph{unclear}.
        \item We showed a list of human-readable labels from the ontology terms that were aligned with the field name.
        \item We asked the expert to discuss each ontology term along the categories described in Table \ref{tab:responsecategories} while voicing their reasoning.
    \end{enumerate}
\end{enumerate}

%---------------- INTERVIEW RESULTS ----------------%
\subsection{Expert Panel Results}
\label{sec:surveyresults}

\begin{figure}[t]
\centering
\frame{\includegraphics[scale=0.43]{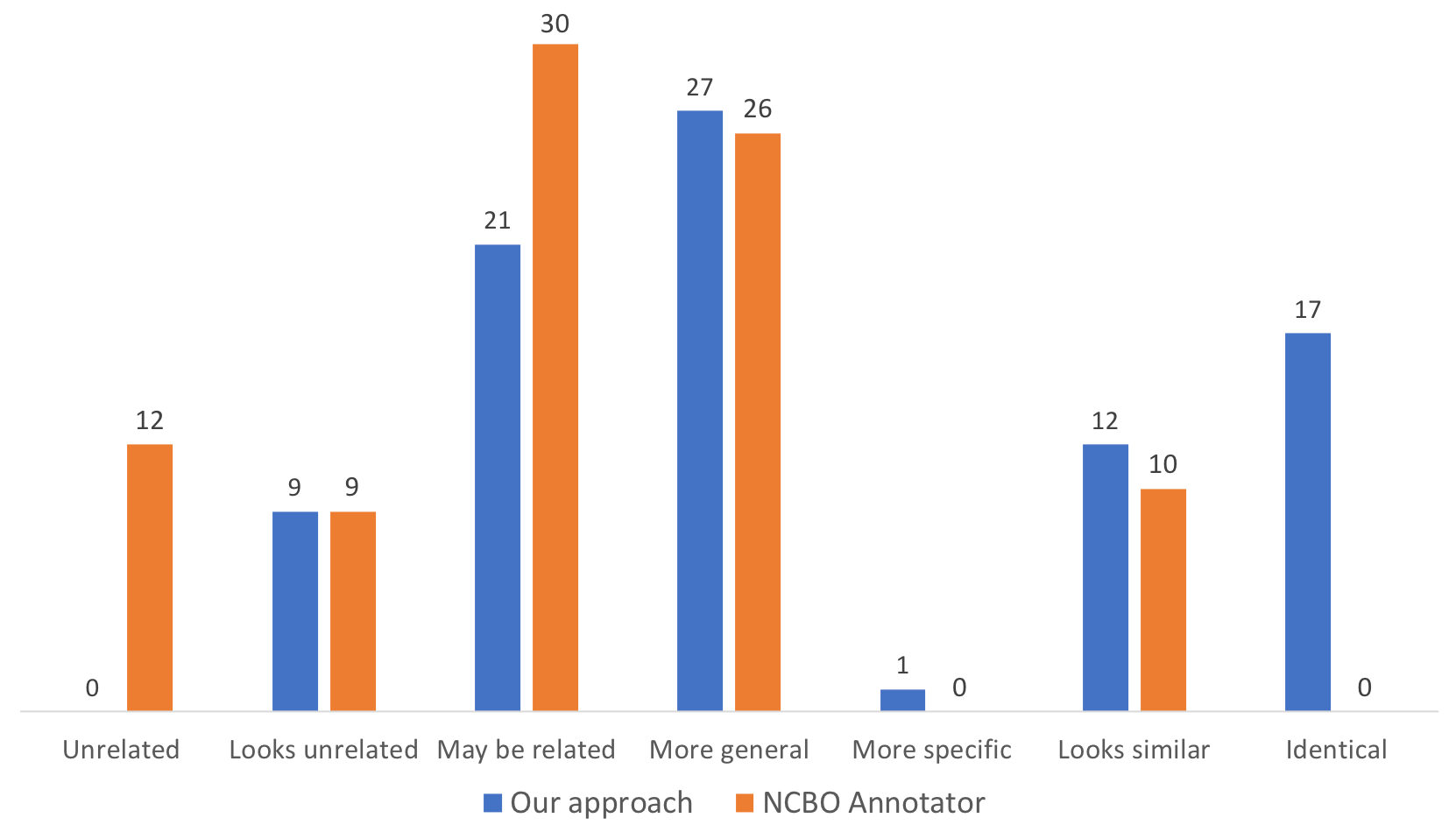}}
\caption{\textbf{Results of expert panel}. Number of alignments that were considered to be in each of the categories along the \emph{x}-axis, for both our method and the NCBO Annotator.}
\label{fig:surveyresultsfigure}
\vspace{-4mm}
\end{figure}

The results of testing \emph{H1}, shown in Figure \ref{fig:surveyresultsfigure}, suggest that many of the alignments found by our approach were of very high quality---in the \emph{identical} category. On the other hand, none of the alignments found by the NCBO Annotator were considered identical to the metadata field names. Among the alignments found by our method, none was considered \emph{unrelated} to the corresponding field name, while the NCBO Annotator suggested 12 such \emph{unrelated} alignments. For example, the NCBO Annotator suggested the term \emph{mapped} for the field name \emph{mapped reference genome} (Q2).

Both methods yielded 9 alignments that \emph{looked unrelated} to the field names. One example of these is the ontology term \emph{cardiovascular function} for the field name \emph{cardiovascular failure} (Q6), as categorized by one expert. On the other hand, this expert considered the ontology term \textit{cardiovascular} (given by the NCBO Annotator) to be \emph{similar} to \textit{cardiovascular failure}. One school of thought may consider \emph{cardiovascular failure} to be \textit{more specific} than \emph{cardiovascular function}. In this case, the expert reasoning could also be that \emph{cardiovascular function} indicates ``heart is functioning well'' and \emph{cardiovascular failure} indicates ``heart is not functioning well''. In the future, we will show experts a small snippet of all descendants of the aligned ontology term (with definition) for more context. 
%An ideal method will recommend an ontology term \emph{Cardiovascular function test abnormal} for the metadata field name \emph{cardiovascular failure}.

The alignments found by the NCBO Annotator were often categorized as \emph{more general} than the metadata field names. For example, for the field name \emph{tissue source} (Q3) the NCBO Annotator suggested \emph{tissue} and \emph{source}, while our method suggested \emph{tissue source site} and \emph{source organ}. 
As expected, we found that when the metadata field names were syntactically identical or very similar to the labels of ontology terms, experts categorized them as \emph{identical}. For example, experts categorized \emph{depth of sample} as either identical or similar to the field name \emph{sample depth m}. One expert observed that \emph{depth of sample} is more generic, since it is missing the specification of the unit of measurement.

Interestingly, when answering Q8, two experts thought the meaning of the metadata field name \emph{patientnumber} was clear---they saw the first ontology term recommended, which happened to be ``patient number'' and they said it was identical. When they saw the third term recommended, which was ``number of patients'', both experts revised their interpretation of the meaning of the field name. One expert lowered the alignment of both options to ``similar'', while the other expert admitted that the third option could be just as ``identical'' as the first option to align with the field name.

%To answer Q2, Participant 4 attempted to find a definition of ``mapped reference genome'' on the Web, or mentions of the field name in publications. The participant was unable to find an explicit definition of the whole field name, or to understand precisely what it meant given the contexts in which it is used (even though the participant understood the meaning of ``reference genome''). There is no one ontology term that contains the field name in its label, and so none of the alignments given are identical matches.

The results of testing \emph{H2} demonstrate that most of the alignments found by our method were very precise, and categorized as either \emph{similar} or \emph{identical} by participants. The study showed that 39 (43\%) ontology term alignments were considered \emph{identical} matches to the metadata field names; 23 (26\%) alignments were considered \emph{similar}; 7 (8\%) alignments \emph{more specific} than the field name; 10 (11\%) alignments \emph{more general}; 8 alignments that \emph{may be related}; 1 alignment that \emph{looked unrelated}; and 2 alignments that were considered \emph{unrelated}.

%---------------- CONCLUSIONS ----------------%
\section{Conclusions}
\label{sec:conclusions}

We developed a new method based on clustering and embeddings to align arbitrary strings with ontology terms. We applied our method to align metadata field names from NCBI BioSample metadata with terms from ontologies in the BioPortal repository. In our experiments we determined that clustering metadata field names using affinity propagation according to the Jaro-Winkler distance metric was the most suitable combination for our corpus. Using this combination as the basis of our alignment method, we were able to find high quality alignments between all metadata field names and at least one ontology term. Unlike existing ontology alignment or string similarity methods, we compute semantic similarity using a combination of background knowledge derived from biomedical publication abstracts and ontology term descriptions, and we can efficiently align strings against a corpus of 10 million terms from 675 ontologies. 

We carried out a comparative study between our method and the NCBO Annotator, in which we discovered that our method found many more alignments overall. We also compared the two methods as part of an expert panel that we conducted via semi-structured interviews. We discovered that our method was able to find highly precise alignments even between strings that, while syntactically not the same, described the same thing. On the other hand, none of the alignments found by the NCBO Annotator were considered identical to the metadata field names. The overall results of our expert panel are illustrative of the efficacy of our method, and its potential for applications. In this work we tuned and applied our method to biomedical metadata, although in principle our approach can be applied to any domain that has a sufficiently rich source of textual knowledge and ontologies to generate embeddings from.

Our experiments show that our method is a suitable solution to align biomedical metadata with ontology terms. Aligning and replacing legacy field names in a metadata repository (or multiple repositories) with ontology terms can substantially improve the searchability of the metadata, and thus the discoverability of the associated data. The applications of our method are twofold: it can be used in metadata authoring software to give real-time suggestions for metadata field names, or to give suggestions for field values that should correspond to ontology terms; and it can be used to facilitate metadata cleaning, by equipping scientists with a means to align metadata field names with ontologies.

In future work, we will implement a generic metadata field recommendation service, and subsequently integrate it with the CEDAR metadata authoring tool for recommending (a) field names when users are building metadata templates, and (b) field values when users are filling in metadata templates. We will design a standalone tool that uses our method to: provide a visualization of clusters formed out of input metadata field names; recommend mappings between the field names and ontology terms; give users ranked options for the best terms for specific fields; and generate a new enriched dataset that materializes the selected field name mappings by replacing legacy field names with ontology terms. 

\vspace{-3mm}
\subsubsection{Acknowledgments}
This work is supported by grant U54 AI117925 awarded by the U.S. National Institute of Allergy and Infectious Diseases (NIAID) through funds provided by the Big Data to Knowledge (BD2K) initiative.  BioPortal has been supported by the NIH Common Fund under grant U54 HG004028.

We thank the experts in our evaluation panel: John Graybeal, Josef Hardi, Marcos Mart\'inez-Romero, and Csongor Nyulas (all of whom from the Center for Biomedical Informatics Research at Stanford University), for their participation.

%\subsubsection{Limitations}
%The embeddings we generated for ontology terms were based solely on their human-readable labels represented via \emph{rdfs:label} and \emph{skos:prefLabel}. In the future, we will investigate taking into consideration the synonyms of ontology terms potentially represented using custom annotation properties, as well as alternative labels represented, for example, using \emph{skos:altLabel}.

\bibliographystyle{splncs04}
\bibliography{references-manual}

\begin{thebibliography}{10}
\providecommand{\url}[1]{\texttt{#1}}
\providecommand{\urlprefix}{URL }
\providecommand{\doi}[1]{https://doi.org/#1}

\bibitem{Barrett2012BioProjectMetadata}
Barrett, T., et~al.: {BioProject and BioSample databases at NCBI: facilitating
  capture and organization of metadata}. Nucleic Acids Research  \textbf{40},
  D57--D63 (2012)

\bibitem{ester1996density}
Ester, M., et~al.: A density-based algorithm for discovering clusters in large
  spatial databases with noise. In: Conf.\ on Knowledge Discovery and Data
  Mining (1996)

\bibitem{Frey2007ClusteringPoints}
Frey, B.J., Dueck, D.: {Clustering by passing messages between data points}.
  Science  \textbf{315}(5814),  972--976 (2007)

\bibitem{goldberg2014word2vec}
Goldberg, Y., Levy, O.: word2vec explained: deriving {Mikolov} et al.'s
  negative-sampling word-embedding method. arXiv preprint arXiv:1402.3722
  (2014)

\bibitem{Goncalves2018TheExperiments}
Gon{\c{c}}alves, R.S., Musen, M.A.: The variable quality of metadata about
  biological samples used in biomedical experiments. Scientific Data
  \textbf{6},  190021 (2018)

\bibitem{jimenez2011logmap}
Jim{\'e}nez-Ruiz, E., Grau, B.C.: Logmap: Logic-based and scalable ontology
  matching. In: International Semantic Web Conference. pp. 273--288. Springer
  (2011)

\bibitem{Jonquet2009-ds}
Jonquet, C., et~al.: {NCBO Annotator: Semantic Annotation of Biomedical Data}.
  In: International Semantic Web Conference (2009)

\bibitem{kamdaremperical}
Kamdar, M.R., et~al.: An empirical meta-analysis of the life sciences (linked?)
  open data cloud (2018), accessible at:
  \url{http://onto-apps.stanford.edu/lslodminer}

\bibitem{koster2007parsing}
Koster, C., Seutter, M., Seibert, O.: {Parsing the Medline Corpus}. In: Recent
  Advances in Natural Language Processing (2007)

\bibitem{lin2015learning}
Lin, Y., et~al.: Learning entity and relation embeddings for knowledge graph
  completion. In: AAAI Conference on Artificial Intelligence (2015)

\bibitem{mcinnes2017hdbscan}
McInnes, L., Healy, J., Astels, S.: {HDBSCAN: Hierarchical density based
  clustering}. The Journal of Open Source Software  \textbf{2}(11), ~205 (2017)

\bibitem{Noy2009BioPortal:Mouse}
Noy, N.F., et~al.: {BioPortal: ontologies and integrated data resources at the
  click of a mouse}. Nucleic Acids Research  \textbf{37},  W170--W173 (2009)

\bibitem{passos2014lexicon}
Passos, A., Kumar, V., McCallum, A.: Lexicon infused phrase embeddings for
  named entity resolution. arXiv preprint arXiv:1404.5367  (2014)

\bibitem{pennington2014glove}
Pennington, J., Socher, R., Manning, C.: {GloVe: Global Vectors for Word
  Representation}. In: Empirical Methods in Natural Language Processing (2014)

\bibitem{percha2018global}
Percha, B., Altman, R.B., Wren, J.: A global network of biomedical
  relationships derived from text. Bioinformatics  \textbf{1}, ~11 (2018)

\bibitem{ristoski2016rdf2vec}
Ristoski, P., Paulheim, H.: {Rdf2vec: RDF graph embeddings for data mining}.
  In: International Semantic Web Conference. pp. 498--514 (2016)

\bibitem{shah2009comparison}
Shah, N.H., et~al.: Comparison of concept recognizers for building the open
  biomedical annotator. In: BMC Bioinformatics. vol.~10, p.~S14. BioMed Central
  (2009)

\bibitem{smaili2018opa2vec}
Smaili, F.Z., Gao, X., Hoehndorf, R.: Opa2vec: combining formal and informal
  content of biomedical ontologies to improve similarity-based prediction.
  arXiv preprint arXiv:1804.10922  (2018)

\bibitem{socher2013reasoning}
Socher, R., et~al.: Reasoning with neural tensor networks for knowledge base
  completion. In: Advances in Neural Information Processing Systems (2013)

\bibitem{wang2018comparison}
Wang, Y., Liu, S., Afzal, N., Rastegar-Mojarad, M., Wang, L., Shen, F.,
  Kingsbury, P., Liu, H.: A comparison of word embeddings for the biomedical
  natural language processing. Journal of biomedical informatics  \textbf{87},
  12--20 (2018)

\bibitem{wang2014knowledge}
Wang, Z., Zhang, J., Feng, J., Chen, Z.: Knowledge graph embedding by
  translating on hyperplanes. In: AAAI Conference on Artificial Intelligence
  (2014)

\bibitem{Wilkinson2016}
Wilkinson, M.D., et~al.: {The FAIR Guiding Principles for scientific data
  management and stewardship}. Scientific Data  \textbf{3},  160018 (2016)

\end{thebibliography}

\end{document}